\documentclass[runningheads, orivec]{llncs}
\usepackage[T1]{fontenc}
\usepackage{graphicx}
\usepackage{booktabs}
\usepackage[misc]{ifsym}

\usepackage{amsfonts}
\usepackage{array}
\usepackage{textcomp}
\usepackage{stfloats}
\usepackage{latexsym}
\usepackage{booktabs}
\usepackage[inline]{enumitem}
\usepackage{color, colortbl, xcolor}
\usepackage{soul} 
\usepackage{mathtools}
\usepackage{graphicx, amssymb, mathrsfs, stmaryrd, dsfont}
\usepackage{xparse}
\usepackage{bbold, verbatim}
\usepackage{tcolorbox}

\usepackage{url}

\usepackage{hyperref} 
\usepackage[capitalize,noabbrev]{cleveref}
\usepackage{tikz}
\usetikzlibrary{arrows, positioning,shapes, patterns,shapes.misc}
\usetikzlibrary{shapes, arrows.meta, positioning, calc, backgrounds, fit, shadows}
\usepackage{pgfplots}
\usepackage{pgfplotstable}
\pgfplotsset{compat=newest}
\usepgfplotslibrary{external,units,colorbrewer,groupplots,fillbetween,statistics}
\usepackage{thm-restate}
\usepackage{multirow}
\usepackage{xspace}
\usepackage{mdframed}
\usepackage{siunitx}
\usepackage[abbreviations]{foreign}
\usepackage[bottom]{footmisc}
\usepackage{fontawesome}
\usepackage{scalerel} 
\usepackage{subcaption}

\allowdisplaybreaks

\usepackage{acronym}

\newcommand{\metric}{\mu}

\newcommand{\original}{h_\mathrm{verify}} 
\newcommand{\manipulated}{h_\mathrm{audit}} 
\newcommand{\fairness}{\mu} 

\graphicspath{ {figures/} }
\crefname{assumption}{assumption}{assumptions}

\newcommand{\newgroupwidth}[2]
{\expandafter\xdef\csname groupwidth#1\endcsname{#2}}

\begin{document}

\acrodef{honest}{Honest}
\acrodef{ROC}{ROC Mitigation}
\acrodef{OT-L}{Optimal Label Transport}
\acrodef{LinR}{Linear Relaxation}
\acrodef{ThreshOpt}{Threshold Manipulation}
\acrodef{GBDT}{Gradient Boosted Decision Tree}
\acrodef{Log. Reg.}{Logistic Regression}

\acrodef{API}{application programming interface}
\acrodef{ML}{machine learning}
\acrodef{FL}{federated learning \cite{mcmahan2017communication}}
\acrodef{LLM}{large language model}
\acrodef{GNN}{graph neural network}
\acrodef{DePa}{demographic parity}
\acrodef{EqOd}{equalized odds}
\acrodef{EqOp}{equal opportunity}
\acrodef{DiIm}{disparate impact}
\acrodef{DiPr}{differential privacy}
\acrodef{AI Act}{Artificial Intelligence Act}
\acrodef{DSA}{Digital Services Act}
\acrodef{DMA}{Digital Market Act \cite{DMA}}
\acrodef{GDPR}{General Data Protection Regulation \cite{GDPR}}

\title{Leveraging Imperfect Sources to Detect Fairwashing in Black-Box Auditing}
\author{Jade Garcia Bourr\'ee\inst{1} \and
Erwan Le Merrer\inst{1}\and
Beno\^it Rottembourg\inst{2} \and
Gilles Tredan\inst{3}}

\authorrunning{J. Garcia Bourr\'ee et al.}

\institute{Inria, University of Rennes, Rennes, France
\and
Inria, Paris, France
\and
LAAS, CNRS, Toulouse, France}

\maketitle

\begin{abstract}
Algorithmic auditing has become central to platform accountability under frameworks such as the AI Act and the Digital Services Act. In practice, this obligation is discharged through dedicated Audit APIs. This architecture creates a paradox: the entity under scrutiny controls the evaluation interface. A platform facing legal sanctions can serve a compliant surrogate model on its Audit API, while running a discriminatory production system. This deceptive practice is known as fairwashing. Manipulation is undetectable if the auditor relies on only one source.
To address this limitation, we introduce the Two-Source Audit Model (2SAM). This model cross-references the Audit API with an independent trusted stream. The key insight is that the trusted stream does not need to be perfectly aligned with the Audit API. We introduce a consistency proxy, a probabilistic mapping that can reconcile discrepancies between sources. This approach yields three results. First, we quantify the rate of manipulation above which a single-source auditor is blind. Second, we show how proxy quality governs detection power. Third, we provide a closed-form budget condition guaranteeing detection at any target confidence level, closing the blind spot mentioned above.
We validate 2SAM on the UCI Adult dataset, achieving $70\%$ detection power with as few as $127$ cross-verification queries out of a total budget of $750$, using a name-based gender proxy with $94.2\%$ accuracy.
\end{abstract}

\section{Introduction}
\label{sec:intro}

The widespread deployment of machine learning systems in high-stakes decision making has led legislators to mandate external audits. Frameworks such as the \ac{AI Act}\cite{AIAct} and the \ac{DSA}\cite{DSA} now require platforms to demonstrate that their systems do not perpetuate systemic biases, an obligation discharged in practice through dedicated \textit{Audit APIs}. This architecture contains a structural paradox: the very entity being scrutinized is the sole provider of the data used for its evaluation.

A platform facing legal sanctions has a clear incentive to manipulate its Audit API. It can appear compliant while users see a non-compliant interface. This deceptive practice is called fairwashing~\cite{aivodji2019fairwashing}. This is not theoretical. During the 2024 European Parliament elections, researchers found significant discrepancies between TikTok’s official API and its public interface~\cite{TikTokDSA2024}. Meta’s advertising ecosystem showed similar inconsistencies~\cite{bagchi2024social}. Manipulation was detectable only when another source was available for comparison. Without one, the Audit API’s output raised no suspicion.

We formalize this vulnerability and show that cross-source verification is not just desirable but a mathematical necessity. Prior work~\cite{bourree2025robust} formalizes this idea, assuming the auditor has a perfectly aligned reference dataset (an external dataset matching the system under audit)—an assumption rarely met in practice. We address this by introducing the Two-Source Audit Model (2SAM), which detects manipulation via an imperfect consistency proxy, as previously defined. Our contributions are:
\begin{itemize}
    \item \textbf{A quantification of the single-source blind spot(Section~\ref{ss:failure}):} We derive a closed-form expression for the manipulation rate $\gamma_0$ beyond which a single-source auditor cannot distinguish a fairwashed interface from a genuinely fair one, no matter how many queries are issued (\Cref{lem:gamma0}).
    \item \textbf{The 2SAM and its consistency proxy (Section~\ref{ss:2sam}):} We introduce a probabilistic mapping that reconciles heterogeneous sources despite their discrepancies, and shows that platform-influenced calibration of this proxy renders detection impossible (\Cref{th:poisoning}).
    \item \textbf{Detection power as a function of proxy quality (Section~\ref{ss:guarantees}):} We characterize how proxy noise propagates into detection power (\Cref{th:power}), and provide a condition to bypass the blind spot identified above (\Cref{th:mitigation}).
    \item \textbf{Experimental validation (Section~\ref{sec:experiments}):} We validate 2SAM on the UCI Adult dataset, demonstrating $70\%$ detection power with as few as $127$ cross-verification queries out of a total number of $750$, using a name-based gender proxy with $94.2\%$ accuracy. It shows that strong guarantees are achievable at modest cost, even with an imperfect proxy.
\end{itemize}

\section{Related Work}
\label{sec:related}

\subsubsection{Black-box fairness auditing.} Fairness auditing through black-box query access has been formalized as a statistical estimation problem~\cite{dwork2012fairness,feldman2015certifying,yan2022active}, where drawing significant conclusions under a finite query budget is a central challenge~\cite{sandvig2014auditing}. Active auditing strategies have been proposed to improve query efficiency~\cite{yan2022active,maneriker2023online}. However, these approaches assume that the auditor can impose consistency on the queried model by hypothesis, which sidesteps the manipulation problem entirely. More fundamentally, Godinot et al.~\cite{augustin} show that under adversarial manipulation, the difficulty of auditing varies across models in ways that active strategies cannot compensate for.

\subsubsection{Adversarial auditing and fairwashing.} The concept of fairwashing, where a model appears fair only during audits, was introduced by Aivodji et al.~\cite{aivodji2019fairwashing} and developed further by Fukuchi et al.~\cite{fukuchi2020faking}. Constructive responses have been proposed. Garcia Bourrée et al.~\cite{bourree2025robust} show that an auditor with a private, labeled reference dataset can detect manipulation. We build on this two-source insight and relax its key assumption. Our trusted stream need not align perfectly with the Audit API, which lets us cover practical cases that~\cite{bourree2025robust} cannot handle. We also provide a way to obtain the ground-truth labels.

\subsubsection{Proxies for sensitive attributes.} Proxy-based inference of sensitive attributes (e.g., using correlated variables such as geographic location to infer attributes like race or gender) has been studied for model training, fairness evaluation, and audit contexts. The impact of proxy noise (the inaccuracy introduced when proxies imperfectly reflect sensitive attributes) on audit reliability is still understudied. Existing work assumes perfect proxies or studies noise in estimation tasks. We differ on both counts. We use proxies as consistency bridges, not estimators, and provide explicit guarantees on how proxy noise affects detection power.

\subsubsection{Distribution shift and cryptographic approaches.} The 2SAM consistency check is structurally related to two-sample testing and data drift monitoring. Another line of work addresses accountability using cryptographic mechanisms that offer stronger guarantees but require zero-knowledge proofs, which are not currently supported by regulation. The 2SAM relies solely on API queries and provides probabilistic guarantees that meet current regulatory requirements.

\section{Methodology}
\label{sec:methodology}

This section develops our approach along four steps. We first define the auditing game and its actors~(\Cref{ss:model}). This model is standard and serves as the foundation for our analysis. We then show the preliminary negative result that single-source auditing is structurally insufficient~(\Cref{ss:failure}): \Cref{th:gap} establishes in our formalism that fairwashing is undetectable without an external reference, and \Cref{lem:gamma0} quantifies the resulting blind spot $\gamma_0$. These two results motivate our main contribution: the Two-Source Audit Model (2SAM), which introduces a \textit{trusted second stream}, formalizes the audit protocol, and defines the consistency proxy $\phi$~(\Cref{ss:2sam}). Finally, \Cref{ss:guarantees} derives the statistical guarantees of the 2SAM; \Cref{th:power} and \Cref{th:mitigation} are our core theoretical contributions: they characterize detection power as a function of proxy quality $p_\phi$, and provide an explicit budget condition that closes the blind spot identified in~\Cref{ss:failure}.

\subsection{The Auditing Game: Formal Model}
\label{ss:model}

We now introduce the formal model underlying our analysis.

\subsubsection{Actors and spaces.}  We consider two actors: a platform and an auditor. Upon regulatory request, the platform exposes a dedicated Audit API: a query interface to access a decision model $h_\mathrm{audit} : \mathcal{Z} \to \mathcal{Y}$, where $\mathcal{Z} = \mathcal{X} \times \mathcal{S}$ is the input space consisting of non-sensitive features $\mathcal{X} \subseteq \mathbb{R}^d$ and a binary sensitive feature $\mathcal{S} \in \{0,1\}$, and $\mathcal{Y} = \{0,1\}$ is the binary output space. On the Audit API, the auditor controls the inputs $Q = \{z_i\}_{i=1}^{N\mathrm{audit}}$ with $z = (x, s) \in \mathcal{Z}$ drawn from a distribution $\mathcal{D}$, and observes the corresponding output labels.

\subsubsection{Fairness metric and estimation error.} The auditor seeks to verify whether the platform satisfies $\mu(h_\mathrm{audit}, \mathcal{D}) \geq \tau$, with $\tau$ the compliance threshold. Throughout this paper, we instantiate $\mu$ as the \textit{Disparate Impact} (DI) metric~\cite{feldman2015certifying}, defined as the ratio of acceptance rates between the unprivileged ($S=0$) and privileged ($S=1$) groups:
\begin{equation}\label{eq:DI}
\mu(h_\mathrm{audit}) = \frac{\mathbb{P}{z \sim \mathcal{D}}(h\mathrm{audit}(z) = 1 \mid S = 0)}
{\mathbb{P}{z \sim \mathcal{D}}(h\mathrm{audit}(z) = 1 \mid S = 1)},
\end{equation}
with compliance threshold $\tau = 80\%$, following the EEOC 80\% rule~\cite{EEOC}. Because the auditor issues a finite number of queries $N_\mathrm{audit}$, the estimated metric $\hat{\mu}(h_\mathrm{audit})$ is subject to sampling variability. At significance level $\alpha$, the margin of error over $N_\mathrm{audit}$ queries is:
\begin{equation}\label{eq:margin}
\epsilon = z_\alpha \frac{\sigma}{\sqrt{N_\mathrm{audit}}},
\end{equation}
where $z_\alpha$ is the normal quantile and $\sigma/\sqrt{N_\mathrm{audit}}$ is the standard error. The auditor concludes that the platform is \textit{compliant} if $\hat{\mu}(h_\mathrm{audit}) \geq \tau$, and \textit{non-compliant} otherwise.

\subsubsection{Manipulation detection.} So far, the auditor’s goal has been framed as estimation: deciding whether $\mu(h_\mathrm{audit}, \mathcal{D}) \geq \tau$. A more adversarial objective arises when the platform may strategically manipulate the Audit API. In this case, the auditor must additionally determine whether the interface it queries faithfully reflects the platform’s true production model $h_\mathrm{verify}$, or whether it has been replaced by a compliant surrogate $h_\mathrm{audit} \neq h_\mathrm{verify}$.

The core challenge is to quantify the expected discrepancy between the two models: how much their outputs can differ, and under what conditions this difference becomes statistically distinguishable from noise. This gap is the detection target of our framework.

We call this second objective manipulation detection. It is logically distinct from fairness estimation: a platform may be simultaneously fair and honest, fair but dishonest, or unfair and dishonest.

\subsection{The Failure of Single-Source Auditing}
\label{ss:failure}

The fundamental weakness of single-source audits lies in the incentive misalignment created by regulations such as the AI Act and the DSA: the platform controls both the system being evaluated and the interface used to evaluate it.

\subsubsection{Adversarial reduction.} The auditor controls the inputs $z = (x, s)$ submitted to the Audit API and observes only the outputs. Consequently, any internal transformation the platform applies (\textit{e.g.,} weight fine-tuning~\cite{hardt2016equality}, reject-option classification~\cite{kamiran2012decision}, post-processing~\cite{jiangWassersteinFairClassification2020,lohausTooRelaxedBe2020}) is observable solely through its effect on output labels. Without loss of generality, every manipulation strategy therefore reduces to a probabilistic \textit{label-flipping}. This reduction is key to the generality of our results: the guarantees derived below hold against any manipulation mechanism, not just label-flipping.

\begin{lemma}[Detectability Gap]\label{th:gap}
In a strict black-box setting where the auditor only has access to a single Audit API, any manipulation that satisfies the fairness threshold is undetectable.
\end{lemma}

\begin{proof} The platform knows the queries $Q$ and the function $\mu$ that the auditor estimates. If the platform’s model $h_\mathrm{verify}$ is already in compliance ($\metric(h_\mathrm{verify}, Q) \geq \tau$), then the platform returns the non-manipulated $h_\mathrm{verify}$ to the auditor. If the property is violated ($\metric(h_\mathrm{verify}, Q) < \tau$), the platform risks being sanctioned. Hence, there is an incentive to find another model $h_\mathrm{audit}$ that satisfies the property.
First, assume that no such $h_\mathrm{audit}$ exists: this means that there exists a query set $Q$ for which every possible answer leads to a violation of the property. In other words, all platforms are guilty given $Q$, and the check is trivial. In this scenario, the audit cannot conclude that a platform is fair.
Second, consider that the platform finds a model $h_\mathrm{audit}$ that satisfies the property ($\metric(h_\mathrm{audit}, Q)\geq \tau$). The auditor estimates fairness on the distribution given by $h_\mathrm{audit}$ (\textit{i.e.} the fairwashed responses), and not the one given by $h_\mathrm{verif}$. The auditor then erroneously concludes that the platform is compliant because $h_\mathrm{audit}$ is.
Furthermore, fairwashed responses can be made internally consistent, rendering the manipulation indistinguishable from a naturally fair model~\cite{fukuchi2020faking}. The audit, therefore, concludes on the compliance of the interface, not the platform.
\end{proof}

\subsubsection{Use case.}
We assume that the platform applies \textit{positive discrimination} by flipping negative labels to positive for the unprivileged group. The manipulation rate $\gamma$ is defined as the fraction of queries on which $h_\mathrm{audit}$ and $h_\mathrm{verify}$ disagree for the unprivileged group:
\begin{equation}
\gamma = \frac{|\{z : h_\mathrm{verify}(z)=1,
h_\mathrm{audit}(z)=0, S=0\}|}
{|\{z : h_\mathrm{audit}(z)=0, S=0\}|}.
\end{equation}

\begin{restatable}{theorem}{deceptionbound}\label{lem:gamma0}
At significance level $\alpha$, the auditor using $N_\mathrm{audit}$ queries on the Audit API is blind to manipulations at rate $\gamma \geq \gamma_0$, where:
\begin{equation}\label{eq:gamma0}
\gamma_0 = \frac{z_\alpha \sigma}
{\sqrt{N_\mathrm{audit}}\left(\tau -
\frac{1}{Y_\mathrm{audit}}\right)},
\end{equation}

with $Y_\mathrm{audit} = \mathbb{P}(h_\mathrm{audit}(z)=1 \mid S=1)$ and $\sigma/\sqrt{N_\mathrm{audit}}$ the standard error of Eq.~\eqref{eq:margin}. The proof is deferred to Appendix~\ref{a:8401}.
\end{restatable}

\paragraph{Takeaway.} The blind spot $\gamma_0$ is not a limitation of the auditor’s strategy but a fundamental property of the single-source setting: no query budget can compensate for the absence of an external reference. This motivates the introduction of a trusted stream in Section~\ref{ss:2sam}.

\subsection{The Two-Source Audit Model}
\label{ss:2sam}

\subsubsection{The need for a second source.} \Cref{th:gap} shows in our formalism that without an external reference, fairwashing is structurally undetectable: the auditor cannot distinguish a manipulated interface from a genuinely fair one on the basis of Audit API outputs alone. This impossibility is not a limitation of the auditor’s strategy but a fundamental property of the single-source setting. It motivates introducing an external reference that the platform does not control. We formalize this external source as a trusted stream $s_T$: a data source whose labels reflect the platform’s true behavior and are not subject to strategic manipulation. The auditor’s task then becomes to verify the \textit{consistency} between the Audit API source $s_A$ and $s_T$.

\subsubsection{Protocol.} The 2SAM auditing game proceeds sequentially in six steps.

\begin{enumerate}
    \item \textbf{Audit API construction.} Upon regulatory request, the platform creates and exposes a dedicated Audit API. This step is mandated by frameworks such as the DSA and AI Act: the platform cannot refuse to provide auditor access.
    \item \textbf{Query submission.} The auditor submits $N_\mathrm{audit}$ queries $Q$ to the Audit API.
    \item \textbf{Platform response (manipulation or not).} Upon receiving the auditor’s queries, the platform chooses whether to serve its true production model $h_\mathrm{audit} = h_\mathrm{verify}$ or a fairwashed surrogate $h_\mathrm{audit} \neq h_\mathrm{verify}$. This substitution is triggered in a single, consistent step across all incoming queries.
    \item \textbf{Label collection.} The auditor collects the output labels ${h_\mathrm{audit}(z_i)}{i=1}^{N\mathrm{audit}}$ from the Audit API.
    \item \textbf{Cross-verification.} The auditor collects $N_\mathrm{verify}$ observations from the trusted stream $s_T$ and applies the consistency proxy $\phi$ (defined below) to map each trusted observation to a predicted Audit API observation. The inconsistency count $I = |\{i : \phi(z_{T,i}, y_{T,i}) \neq (z_{A,i}, h_\mathrm{audit}(z_{A,i})\}|$ is computed over the $N_\mathrm{verify}$ cross-verified pairs.
    \item \textbf{Verdict.} The auditor runs two independent tests. A platform is flagged as \textit{Unfair} if it fails the fairness test ($\hat{\mu}(h_\mathrm{audit}) < \tau$) and as \textit{Manipulative} if it fails the consistency test ($I \geq I_0$), where $I_0$ is the rejection threshold derived in \Cref{ss:guarantees}. Otherwise the platform is considered fair ($\hat{\mu}(h_\mathrm{audit}) \geq \tau$) and non-manipulative ($I < I_0$).
\end{enumerate}

\paragraph{Regulatory insight.} This protocol maps directly onto the obligations of the DSA and AI Act. The verdict structure makes an operationally important distinction that current regulatory texts often mix: \textit{unfairness} (the system discriminates) and \textit{fairwashing} (the system discriminates but conceals it through the Audit API) require different legal responses. Separating the two tests provides regulators with actionable, legally distinguishable outcomes.

\subsubsection{The consistency proxy.} In general, $s_A$ and $s_T$ will not share the same feature space or encoding. What constitutes a valid trusted stream depends on the context. In the setting of~\cite{bourree2025robust}, $s_T$ is a privately labeled reference dataset held by the auditor, a strong assumption that provides a perfect baseline but is costly to acquire. Our contribution is to relax this assumption: we show in \Cref{ss:guarantees} that $s_T$ need not be perfectly aligned with $s_A$ to enable detection, as long as the degree of misalignment can be characterized.

We first consider the idealized case of a perfect mapping between the two sources: every observation from $s_T$ can be matched unambiguously to a corresponding Audit API query-call (as was the case for the historical $\mathds{X}$ API~\cite{twitter}). Here, any inconsistency is necessarily caused by platform manipulation; a single inconsistency suffices to flag fairwashing.

In practice, such a perfect mapping is rarely available: the two sources may differ in feature space, granularity, stochasticity, or encoding, so that even an honest platform produces some apparent inconsistencies. To handle this realistic setting, we introduce a consistency proxy $\phi$ that reconciles the two sources despite their discrepancies:

\begin{definition}[Proxy]\label{def:proxy}
 A proxy is a probabilistic mapping
 \begin{equation*}
 \phi: \mathcal{Z}_T \times \{0,1\} \;\longrightarrow\;
 \mathcal{Z}_A \times \{0,1\},\qquad
 (z_T, y_T) \;\longmapsto\; (z_A, y_A)
 \quad \text{w.p.}\ p_{z_T, y_T}.
 \end{equation*}
 \end{definition}

The quality parameter $p_\phi \in [0,1]$ denotes the average accuracy of $\phi$ on an honest platform. It mathematically decouples measurement noise ($q_\phi = 1 - p_\phi$) from intentional manipulation ($\gamma$). When $p_\phi = 1$, we recover the perfect case: every inconsistency signals manipulation. When $p_\phi < 1$, some inconsistencies arise from proxy noise alone, which motivates the hypothesis test in \Cref{ss:guarantees}.

\subsubsection{Proxy independence.}\label{hyp:proxy} For the consistency check to be valid, $p_\phi$ must be estimated from external sources (\textit{e.g.,} national censuses, demographic surveys, or journalistic investigations), never from platform-provided metadata. This independence condition is justified through the following result.

\begin{lemma}[Proxy Poisoning]\label{th:poisoning}
In a strict black-box setting where the auditor relies on a proxy quality parameter $p_\phi$ derived from the platform’s own declarations, any manipulation satisfying the fairness threshold remains statistically undetectable.
\end{lemma}

\begin{proof}[Proof Sketch] The proof is deferred to Appendix~\ref{a:84012}.
If the platform influences the estimation of $p_\phi$, it can align the expected noise $q_\phi$ with its manipulation rate $\gamma$, ensuring that the empirical inconsistency count $I$ remains low and the consistency check is neutralized.
\end{proof}

\paragraph{Regulatory insight.} \Cref{th:poisoning} has a direct regulatory implication: platforms must never be allowed to provide or influence the metadata used to calibrate the auditor’s proxy. This is the algorithmic analogue of the auditor independence principle in financial auditing.

\subsection{Detection Guarantees}
\label{ss:guarantees}

We now derive the statistical guarantees of the 2SAM against fairwashing manipulation of the Audit API. The key question is: \textbf{when $p_\phi < 1$, how many inconsistencies are needed to distinguish proxy noise from intentional manipulation?}

\subsubsection{Hypothesis test.}
Under $H_0$ (honest platform, $h_\mathrm{audit} = h_\mathrm{verify}$), inconsistencies arise only from proxy noise: $I \sim \mathrm{Binomial}(N_\mathrm{verify}, q_\phi)$. The rejection threshold controlling the Type~I error (false accusation) at level $\alpha$ is:
 \begin{equation}\label{eq:I0}
 I_0 = \min\!\left\{n : \sum_{i=0}^{n}
 \binom{N_\mathrm{verify}}{i}
 q_\phi^i\, p_\phi^{N_\mathrm{verify}-i} \geq 1-\alpha \right\}.
 \end{equation}
Under $H_1$ (fairwashing at rate $\gamma$), the inconsistency probability increases to $q_\phi^\gamma = \gamma p_\phi + (1-\gamma)q_\phi$: an inconsistency occurs either when the proxy correctly identifies a manipulated label (prob.\ $\gamma p_\phi$) or incorrectly predicts an honest one (prob.\ $(1-\gamma)q_\phi$).

\begin{theorem}[Detection Power]\label{th:power}
With significance level $\alpha$ and $N_\mathrm{verify}$ queries, the probability that the auditor detects manipulation at rate $\gamma$ is:
\begin{equation}\label{eq:power}
p_\mathrm{manip}(\gamma) = 1 - F_{\mathcal{B}(N_\mathrm{verify},
q_\phi^\gamma)}(I_0),
\end{equation}
where $F_{\mathcal{B}}$ is the binomial CDF and $q_\phi^\gamma = \gamma p_\phi + (1-\gamma)q_\phi$.
\end{theorem}

\begin{proof}
Under $H_1$, $I \sim \mathrm{Binomial}(N_\mathrm{verify}, q_\phi^\gamma)$. The detection probability is the tail probability beyond the rejection threshold: $p_\mathrm{manip}(\gamma) = 1 - F_{\mathcal{B}(N_\mathrm{verify}, q_\phi^\gamma)}(I_0)$.
\end{proof}

\paragraph{Takeaway.} The probability that the auditor detects manipulation $p_\mathrm{manip}$ is increasing in both $\gamma$ and $p_\phi$: a more accurate proxy detects weaker manipulations.

\subsubsection{Use case.} In the use case of the platform using positive-discrimination fairwashing, we can derive the mitigation guarantee under which the auditor succeeds the audit. To do so, we assume that the auditor uses $N_\mathrm{audit}$ queries to estimate the fairness on the Audit API and verifies $N_\mathrm{verify}$ queries through its trusted source.

 \begin{theorem}[2SAM Mitigation Guarantee]\label{th:mitigation}
 A Two-Source Audit succeeds with confidence $1-\alpha$ against positive-discrimination fairwashing if the budget $(N_\mathrm{audit}, N_\mathrm{verify})$ satisfies:
 \begin{equation}
 F_{\mathcal{B}(N_\mathrm{verify},\,
 q_\phi^{\gamma_0})}(I_0) \leq \alpha,
 \end{equation}
with $\gamma_0 = \frac{z_\alpha\, \sigma}
 {\sqrt{N_\mathrm{audit}}\!\left(\tau -
 \frac{1}{Y_\mathrm{audit}}\right)}$ and $I_0 = \min\!\left\{n : \sum_{i=0}^{n}
 \binom{N_\mathrm{verify}}{i}
 q_\phi^i\, p_\phi^{N_\mathrm{verify}-i} \geq 1-\alpha \right\}$.
 \end{theorem}

\begin{proof}
If $\gamma < \gamma_0$: the platform fails the fairness test on the Audit API with probability $\geq 1-\alpha$ by definition of $\gamma_0$. If $\gamma \geq \gamma_0$: the platform may pass the fairness test, but since $p_\mathrm{manip}$ is monotone in $\gamma$, the detection probability is at least $p_\mathrm{manip}(\gamma_0) \geq 1-\alpha$ by the budget condition.
\end{proof}

\subsection{Instantiation: the User Interface as a trusted stream} The model above applies to any trusted stream $s_T$ with a consistency proxy $p_\phi$. A natural and practically available candidate, particularly relevant for regulators operating under the DSA or AI Act, is the platform’s own \textit{User Interface}, the public interface actually encountered by users. It is the source that empirical studies have already used intuitively: comparisons between TikTok’s official API and its public interface during the 2024 European elections~\cite{TikTokDSA2025}, and between Meta’s advertising API and observable user-facing behavior~\cite{bagchi2024social}, revealed discrepancies that the Audit API alone would never have surfaced. The 2SAM provides formal guarantees for precisely this kind of cross-source comparison.

\subsubsection{User Interface Integrity}\label{hyp:integrity} Using the User Interface as $s_T$ requires one additional assumption: the platform must not manipulate the User Interface in a targeted manner against the auditor. Auditors can further strengthen this assumption by applying standard anonymization techniques (IP rotation, behavioral mimicry), thereby raising the cost of auditor identification for the platform.

\subsubsection{Illustrations.} We illustrate the proxy with two examples.

\textit{Example 1: Data Consistency in Social Networks.} Consider an API providing structured access to social media data (e.g., the historical $\mathds{X}$ API \cite{twitter} or the TikTok one \cite{TikTokDSA2024}). Such APIs are theoretically redundant with the platform’s web interface, serving primarily to spare auditors the burden of web scraping. In this baseline case, the auditor expects strict identity between the Audit API and the User interface. The proxy is the identity function:
\begin{equation}
\phi_{\mathds{X}}(z_T, y_T) = (z_T, y_T) \text{ with probability } p_\phi = 1.
\end{equation}
Any discrepancy found through scraping would immediately signal an intentional manipulation in the data served to auditors versus users.

\textit{Example 2 (Recruitment algorithms~\cite{AmazonHiring}).} Hiring tools often face scrutiny for gender or racial bias \cite{AmazonHiring}. In an audit setting, the Audit API might provide explicit fields for protected features (e.g., \textit{Gender}), whereas the User Interface (the resume submission portal) only contains text. The auditor can leverage a proxy $\phi_{gen}$ that estimates gender from first names based on linguistic patterns \cite{barry2000three,basu2022measuring}. The proxy $\phi_\mathrm{gen}$ estimates gender from first names via a linguistic heuristic~\cite{barry2000three}. Names ending in \textit{a, e, i} map to Female ($p=1$); names ending in \textit{h, y} map ambiguously ($p=0.5$); others map to Male ($p=1$). This proxy achieves $p_\phi = 94.2\%$ on French names and serves as our experimental instantiation (\Cref{sec:experiments}).

\section{Experiments}
\label{sec:experiments}

We empirically validate the 2SAM on the UCI Adult dataset~\cite{Dua:2019} (and on COMPAS\cite{angwin2016} in \Cref{a:48611}). As we do not have access to specialized APIs as regulators would, we simulate the auditing game described in \Cref{ss:model}: the User Interface is a logistic regression trained on the full dataset, and the Audit API is its fairwashed counterpart. We demonstrate how the two-source approach closes the structural blind spot of single-source auditing, and characterize the role of proxy quality across three questions:
\textbf{Q1.} How does proxy quality affect detection power? (\Cref{ss:exp:proxy})
\textbf{Q2.} Does the 2SAM detect fairwashing on a use case? (\Cref{ss:exp:detection})
\textbf{Q3.} What is the budget overhead of two-source auditing? (\Cref{ss:exp:pareto})

\subsection{Experimental Setup}
\label{ss:exp:setup}

\subsubsection{Data and model.} We train a logistic regression to predict whether an individual’s income exceeds \$50,000. This production model serves as our ground truth User Interface and exhibits a significant gender bias, with a disparate impact $\mu = 0.24$, failing the regulatory threshold of $\tau = 80\%$.

\subsubsection{Fairwashing simulation.} To simulate a strategic platform, we implement a label-flipping strategy on the Audit API. The platform applies positive discrimination by flipping negative labels to positive for the unprivileged group (Female) at a rate $\gamma$. As discussed in Appendix~\ref{app:fairwashing}, similar results are achievable using other fairwashing methods.

\subsubsection{Proxy family.} All experiments use a name-based gender proxy, leveraging the first names added to each profile (see \Cref{app:repro} for the full name list, taken from \cite{Insee}). We model the proxy family as follows: the proxy knows the true gender for a fraction $p_\phi \in [0,1]$ of individuals without error, and for the remaining fraction $(1-p_\phi)$ it guesses uniformly at random. Formally, for each individual:
 \begin{equation}
 \phi(x_T, y_T) = \begin{cases}
 {gender}_\mathrm{true} & \text{with probability } p_\phi, \\
 \mathrm{Uniform}(\{0,1\}) & \text{with probability } 1 - p_\phi.
 \end{cases}
 \end{equation}
This parametrization yields a proxy family indexed by $p_\phi \in [0.5, 1]$, where $p_\phi = 0.5$ corresponds to a random proxy and $p_\phi = 1$ to a perfect proxy. It allows us to study the effect of proxy quality continuously across experiments to answer \textbf{Q1} and \textbf{Q2}.

A natural instantiation of this family is the linguistic proxy $\phi_\mathrm{gen}$ introduced in Example~2 (\Cref{ss:2sam}), which maps names to gender based on their last letter~\cite{barry2000three}. In our dataset, this proxy achieves $p_\phi = 94.2\%$ and serves as a reference point in all three experiments.

\subsection{Q1 — Effect of Proxy Quality on Detection Power}
\label{ss:exp:proxy}

We first study how proxy quality $p_\phi$ impacts detection power $p_\mathrm{manip}(\gamma)$, as characterized by \Cref{th:power}. We compute $p_\mathrm{manip}$ using Eq.~\eqref{eq:power} with $N_\mathrm{verify} = 127$ and $\alpha = 0.05$.

\begin{figure}[ht]
\centering
\includegraphics[width=0.6\linewidth]{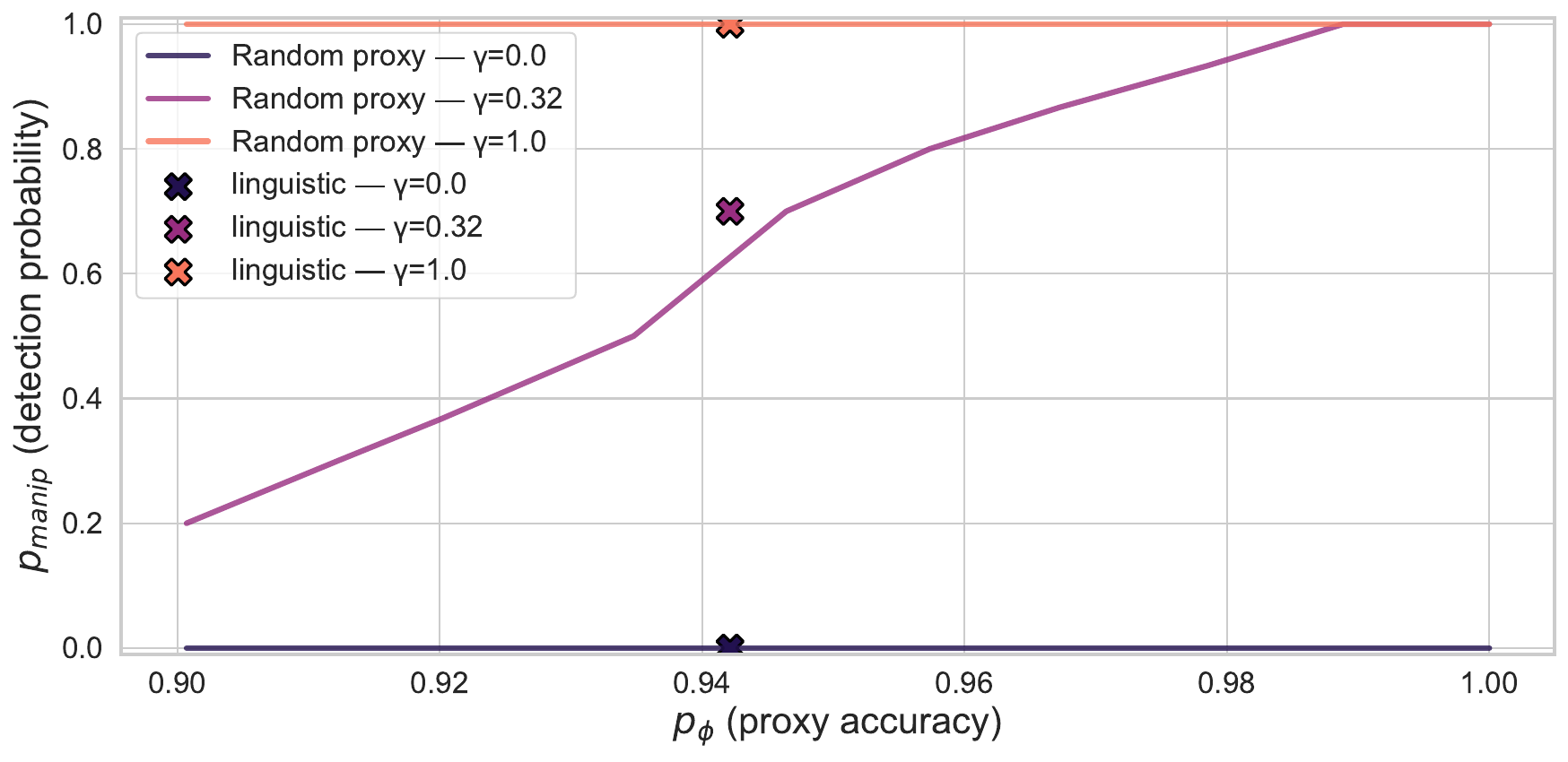}
\caption{Detection power $p_\mathrm{manip}$ as a function of proxy quality $p_\phi$, for several manipulation rates $\gamma$.}
\label{fig:proxy_quality}
\end{figure}

Figure~\ref{fig:proxy_quality} illustrates the detection power $p_{\text{manip}}$ as a function of proxy quality $p_\phi$, for three manipulation rates $\gamma \in \{0.0, 0.32, 1\}$, under two proxy families: the continuous \textit{Noisy} proxy (solid lines) and the \textit{Linguistic} proxy $\phi_{\text{gen}}$ (crosses).

The figure shows a phase transition in the detection power. For $\gamma = 0\%$, $p_{\text{manip}}$ remains at the nominal level $\alpha = 0.05$ across all $p_\phi$, confirming that Type I error is correctly controlled. For $\gamma > 0\%$, $p_{\text{manip}}$ rises steeply with both $p_\phi$ and $\gamma$, in agreement with Theorem~\ref{th:power}. The linguistic proxy $\phi_{\text{gen}}$ ($p_\phi = 94.2\%$) lies in the middle, achieving $p_{\text{manip}} = 70\%$ at $\gamma = 32\%$ and $p_{\text{manip}} = 100\%$ at $\gamma = 100\%$.

\paragraph{Takeaway.} These results establish that imperfect proxies are practically viable for fairwashing detection, but that proxy quality must be verified prior to deployment.

\subsection{Q2 — Fairwashing Detection on a Use Case}
\label{ss:exp:detection}

We now instantiate the full 2SAM workflow on our running example, using the linguistic proxy $\phi_\mathrm{gen}$. This experiment illustrates \Cref{th:gap} (failure of single-source auditing) and \Cref{th:power} (detection by the 2SAM) on the same setting.

\begin{figure}[ht]
\centering
\begin{subfigure}[b]{0.48\linewidth}
\centering
\includegraphics[width=\linewidth]{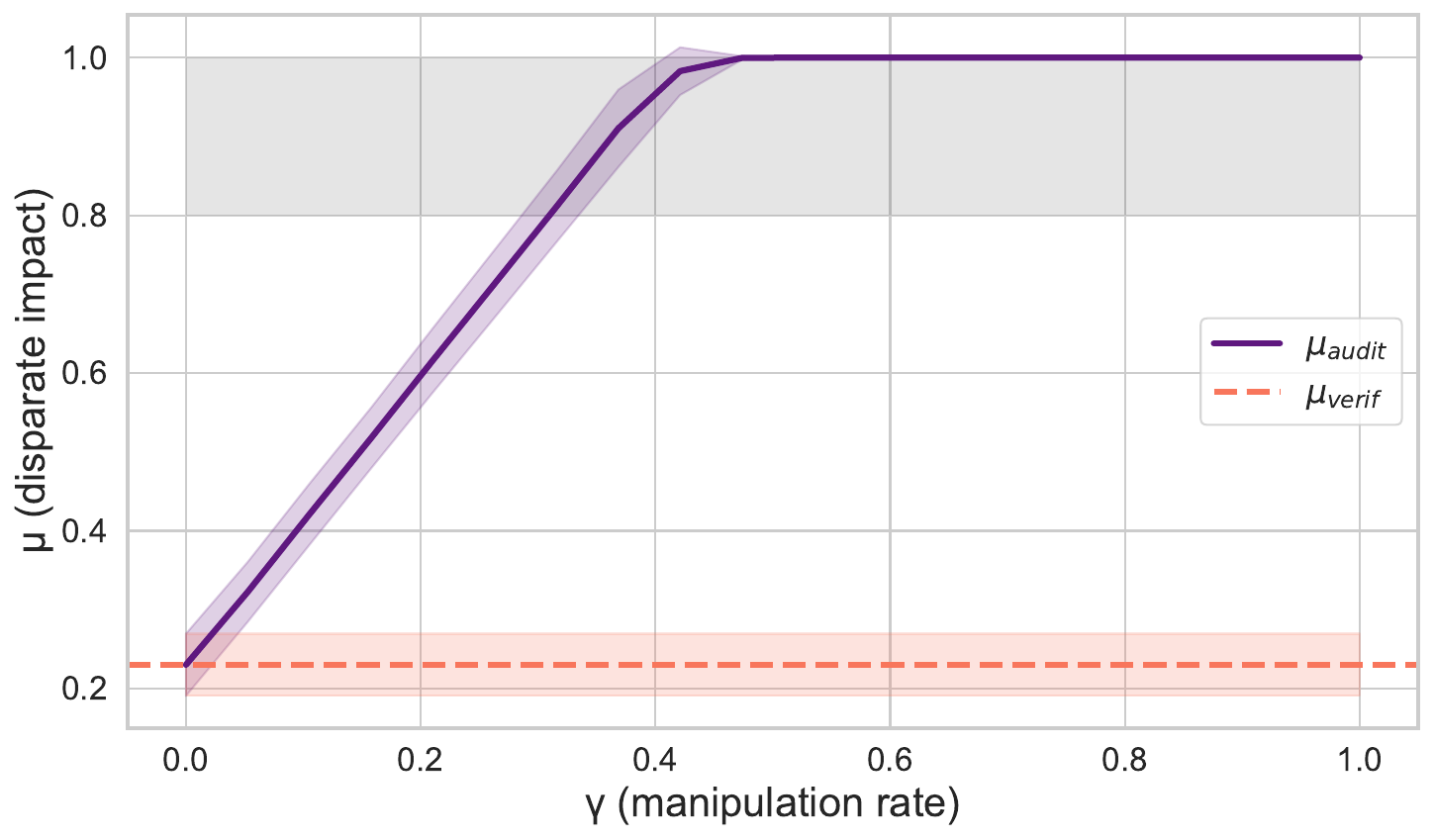}
\caption{Estimated fairness as $\gamma$ increases. }
\label{fig:lies_intermediate}
\end{subfigure}
\hfill
\begin{subfigure}[b]{0.48\linewidth}
\centering
\includegraphics[width=\linewidth]{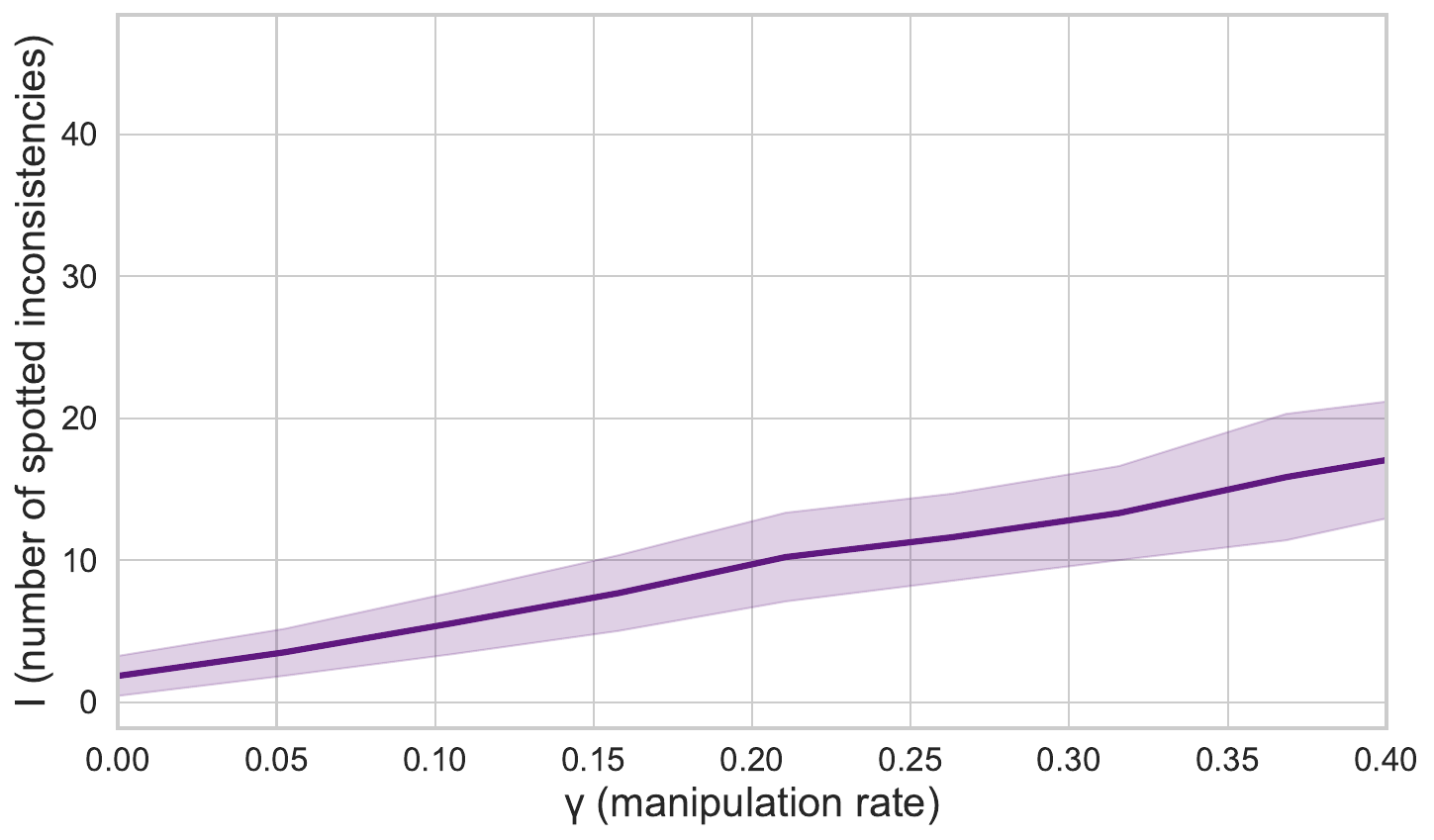}
\caption{Inconsistency count $I$ as $\gamma$ increases.}
\label{fig:inconsistency_count}
\end{subfigure}
\caption{Single-source failure (left) and 2SAM detection (right) under increasing manipulation rate $\gamma$. The gray area marks the fair zone ($\hat{\mu} \geq \tau$). The single-source auditor incorrectly concludes compliance beyond $\gamma_0 = 32\%$.}
\label{fig:detection}
\end{figure}

Figure~\ref{fig:detection} illustrates the Platform’s Dilemma. As the platform increases $\gamma$ toward the deception bound $\gamma_0 = 32\%$, the empirical fairness $\hat{\mu}(h_\mathrm{audit})$ crosses the $\tau = 80\%$ threshold (\Cref{fig:lies_intermediate}). Without an external reference, the single-source auditor concludes that the platform is fair, a false negative that \Cref{th:gap} proves is unavoidable in the single-source setting. However, as $\gamma$ increases, the inconsistency count $I$ rises monotonically: the same label-flipping that drives $\hat{\mu}(h_\mathrm{audit})$ above $\tau$ in Figure~\ref{fig:detection}a simultaneously increases the signal detected by the 2SAM. At some point, $I$ will exceed the rejection threshold $I_0$, and the platform is flagged as manipulative. The platform thus faces an inescapable trade-off: any manipulation sufficient to appear fair to a single-source auditor is detectable by the 2SAM.

\paragraph{Takeaway.} The 2SAM closes exactly the blind spot identified in \Cref{ss:failure}: the same manipulation rate $\gamma_0$ that makes the platform invisible to a single-source auditor makes it detectable to the 2SAM auditor. The proxy noise is absorbed by the hypothesis test without affecting the reliability of the verdict.

\subsection{Q3 — Budget Allocation and Scalability}
\label{ss:exp:pareto}

\begin{figure}[h!]
\centering
\includegraphics[width=0.55\linewidth]{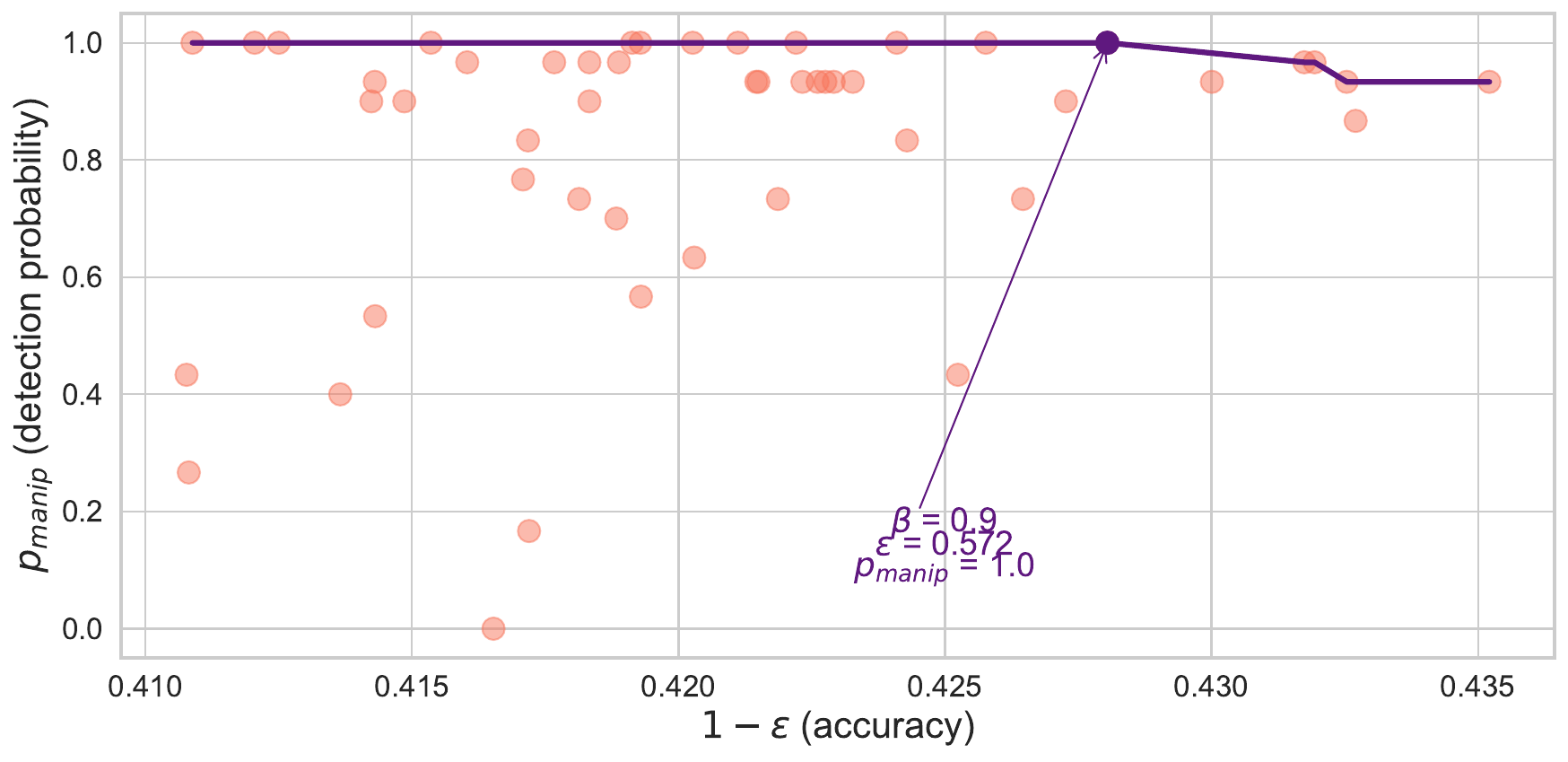}
\caption{Pareto frontier for estimating Disparate Impact while checking for fairwashing under a fixed audit budget ($N = 750$). Each point corresponds to a budget allocation $\beta$.}
\label{fig:pareto}
\end{figure}

So far, we have established that the 2SAM detects manipulation at rate $\gamma_0$ — the minimum rate at which the platform must lie to appear fair. We now ask: at what cost? Figure~\ref{fig:pareto} plots detection power $p_{\text{manip}}$ against estimation accuracy $1 - \epsilon$ for this worst-case manipulation $\gamma_0$, as the auditor varies her budget allocation $\beta \in [0,1]$: a fraction $\beta$ of the total budget $N$ is cross-verified against the User Interface.

The Pareto frontier reveals a clear trade-off: allocating more queries to cross-verification improves detection power at the cost of estimation accuracy. The highlighted point ($\beta = 0.9$, $\epsilon = 57.2\%$) achieves $p_{\text{manip}} = 100\%$ with only around $340$ cross-verification queries out of $N = 750$ — making it statistically impossible for the platform to appear both fair and honest at the same time.

\textbf{Takeaway.} \Cref{th:mitigation} provides a closed-form minimal $N_{\text{verify}}$, and automated proxies eliminate manual labeling costs. Regulators can thus maintain low-intensity monitoring and trigger a high-confidence consistency check only when suspicious patterns emerge. Additional visualizations illustrating the phase-transition structure of detection power as a function of proxy quality, verification budget, and audit/verify ratio $\beta$ are provided in Appendix~\ref{app:additional_figures}.

\subsubsection{Generalization across sensitive attributes and proxy types.} The 2SAM framework is not restricted to gender proxies or income prediction tasks. Appendix~\ref{a:48611} instantiates the same auditing game on the COMPAS dataset using race as the sensitive attribute, demonstrating that the phase-transition structure and detection guarantees of Theorems~\ref{th:power}  and~\ref{th:mitigation} hold across different sensitive attributes, domains, and model families. We acknowledge that proxy construction remains context-dependent: the linguistic proxy $\phi_{\text{gen}}$ is specific to French naming conventions, and practitioners operating in different cultural or demographic contexts should derive proxies from locally appropriate external sources such as national censuses or demographic surveys.

\section{Conclusion}
\label{sec:conclusion}

Algorithmic auditing is only meaningful if the data used for evaluation cannot be manipulated by the entity being evaluated. This condition is not satisfied by current auditing practice: in a single-source setting, any platform facing regulatory scrutiny can engage in undetectable fairwashing by serving a compliant surrogate model on its Audit API. We have quantified the precise manipulation rate $\gamma_0$ beyond which an auditor operating on a single source is blind against positive-discrimination fairwashing.

The Two-Source Audit Model addresses this vulnerability by cross-referencing the Audit API against a trusted stream. Our central finding is that this second source need not be perfect: detection remains statistically guaranteed as long as the consistency proxy exceeds a minimum quality threshold, which is computable before the audit begins and depends only on the target manipulation rate and the auditor’s confidence level. Below this threshold, no amount of additional data can compensate for proxy noise; above it, detection is robust, and the additional query budget is modest, as few as $127$ queries out of $750$ suffice for $70\%$ detection power in our experiments. 

Three implications follow for the design of regulatory audit protocols. First, audit frameworks should mandate auditor access to public-facing interfaces, not only to dedicated Audit APIs: the DSA and AI Act currently focus on the latter while leaving the former unaddressed. Second, platforms must never be allowed to provide or influence the metadata used to calibrate the auditor’s proxy. \Cref{th:poisoning} establishes that platform-influenced calibration renders the entire consistency check void, an analogue of the auditor independence principle in financial regulation. Third, the 2SAM provides regulators with a legally defensible budget formula: given a target detection confidence $1-\alpha$ and a proxy quality estimate $p_\phi$, \Cref{th:mitigation} yields the minimum $N_\mathrm{verify}$ required, removing the arbitrariness from audit resource allocation.

Ultimately, 2SAM highlights that transparency is not a feature that platforms can “provide” through a single interface. It is a property that must be verified through cross-source evidence. Future regulatory standards should emphasize the importance of auditor-led proxy development (with the help of a subject-matter expert and a data scientist) and ensure that platforms cannot obstruct the collection of public-facing data. By moving from a “trust-by-design” to a “verify-by-consistency” model, we provide a foundation for audits that are resilient to deceptive practices.

\bibliography{references}
\bibliographystyle{splncs04}

\onecolumn
\appendix

\section{Proof of Theorem 1}\label{a:8401}
\begin{proof}
With positive discrimination at rate $\gamma$, 
\begin{align*}
P(\manipulated (z) = 1 | S = 0) &= \underbrace{P(\original (z) = 1 | S = 0)}_{\text{already positive output}} + \underbrace{\gamma P(\original (z) = 0 | S = 0)}_{\text{positive discrimination}}\\
&\underset{\text{Unit measure}}{=} P(\original (z) = 1 | S = 0) + \gamma \left( 1 - P(\original (z) = 1 | S = 0) \right)\\
&= P(\original (z) = 1 | S = 0) (1 - \gamma) + \gamma
\end{align*}

And because the decision on the unprivileged group is unchanged:
 $P(\manipulated (z) = 1 | S = 1) = P(\original (z) = 1 | S = 1).$

Thus, by injecting these formulas into the definitions of $\metric$ we obtained:
\begin{align*}
 \metric(\manipulated) &= \frac{P(\manipulated (z) = 1 | S = 0)}{P(\manipulated (z) = 1 | S = 1)}\\
 &= \frac{P(\original (z) = 1 | S = 0) (1 - \gamma) + \gamma}{P(\original (z) = 1 | S = 1)}\\
 &= \metric(\original) (1 - \gamma) + \frac{\gamma}{P(\original (z) = 1 | S = 1)}\\
 &\underset{\text{notation}}{=} \fairness(\original) (1 - \gamma) + \frac{\gamma}{Y_\mathrm{audit}}
\end{align*} 
with $Y_\mathrm{audit} = \mathbb{P}(h_\mathrm{audit}(z)=1 \mid S=1)$.

We recall that the platform wants to appear fair to the auditor ($\hat{\fairness}(\manipulated) \geq \tau$). Using the definition of margin error at confidence level $\alpha$, $\fairness(\manipulated) - \epsilon \leq \hat{\fairness}(\manipulated) \leq \fairness(\manipulated) + \epsilon$. Thus, the platform appears fair only if at least $\fairness(\manipulated) + \epsilon \geq \tau$. Note that this is a necessary but not sufficient condition.

With the previous calculation, it is equivalent to saying that a necessary condition is:
$$\fairness(\original) (1 - \gamma) + \frac{\gamma}{Y_\mathrm{audit}} + \epsilon \geq \tau.$$

As the platform manipulates $\manipulated$ only when $\original$ is not fair (otherwise, $\manipulated = \original$ is sufficient), $$\fairness(\original) \leq \tau.$$

The platform appears fair only if at least $\tau (1 - \gamma) + \frac{\gamma}{Y_\mathrm{audit}} + \epsilon\geq \tau$.

This last equation can be written as well as $\gamma \geq \frac{\epsilon}{\tau - \frac{1}{Y_\mathrm{audit}}}$.

The auditor is thus robust to positive discrimination up to $\frac{\epsilon}{\left(\tau - \frac{1}{Y_\mathrm{audit}}\right)}$ manipulations, which is the expected result.
\end{proof}

\section{Proof of Lemma 2}\label{a:84012}
\begin{proof}
The platform knows $\mu$ and $D$, hence can compute the minimum manipulation 
rate $\gamma$ needed to satisfy $\mu(h_{\text{audit}}) \geq \tau$ via Theorem~1. 
It then estimates the expected inconsistency count under $H_1$:
$$\hat{I} = N_{\text{verify}} \cdot q_\phi^\gamma = N_{\text{verify}} \cdot \left(\gamma p_\phi + (1-\gamma)q_\phi\right).$$
The platform then declares $\tilde{p}_\phi$ such that the auditor's calibrated 
threshold $I_0(\tilde{q}_\phi)$ satisfies $I_0(\tilde{q}_\phi) \geq \hat{I}$, 
i.e.\ such that $\hat{I}$ falls below the rejection region. It 
suffices to find $\tilde{p}_\phi$ such that:
$$\sum_{i=0}^{\hat{I}} \binom{N_{\text{verify}}}{i} \tilde{q}_\phi^i \tilde{p}_\phi^{N_{\text{verify}}-i} \geq 1-\alpha,$$
with $\tilde{q}_\phi = \gamma p_\phi + (1-\gamma)q_\phi$. Since the left-hand 
side is continuous and increasing in $\tilde{q}_\phi$, such a 
$\tilde{p}_\phi = 1 - \tilde{q}_\phi$ always exists. Under this declared value, 
the auditor's threshold absorbs the manipulation signal exactly, and the test 
has power equal to its size $\alpha$ regardless of $N_{\text{verify}}$.
\end{proof}

\section{Additional Figures and Extended Analysis}\label{app:additional_figures}

This appendix presents complementary visualizations that deepen the empirical
understanding of the Two-Source Audit Model (2SAM). These figures illustrate
how proxy quality, manipulation rate, verification budget, and the audit/verify
ratio jointly shape detection power. They also highlight the structural phase
transitions predicted by Theorem~2 and Theorem~3.

\subsection{Threshold Behaviour as a Function of Proxy Quality}

\begin{figure}[h]
    \centering
    \includegraphics[width=.5\linewidth]{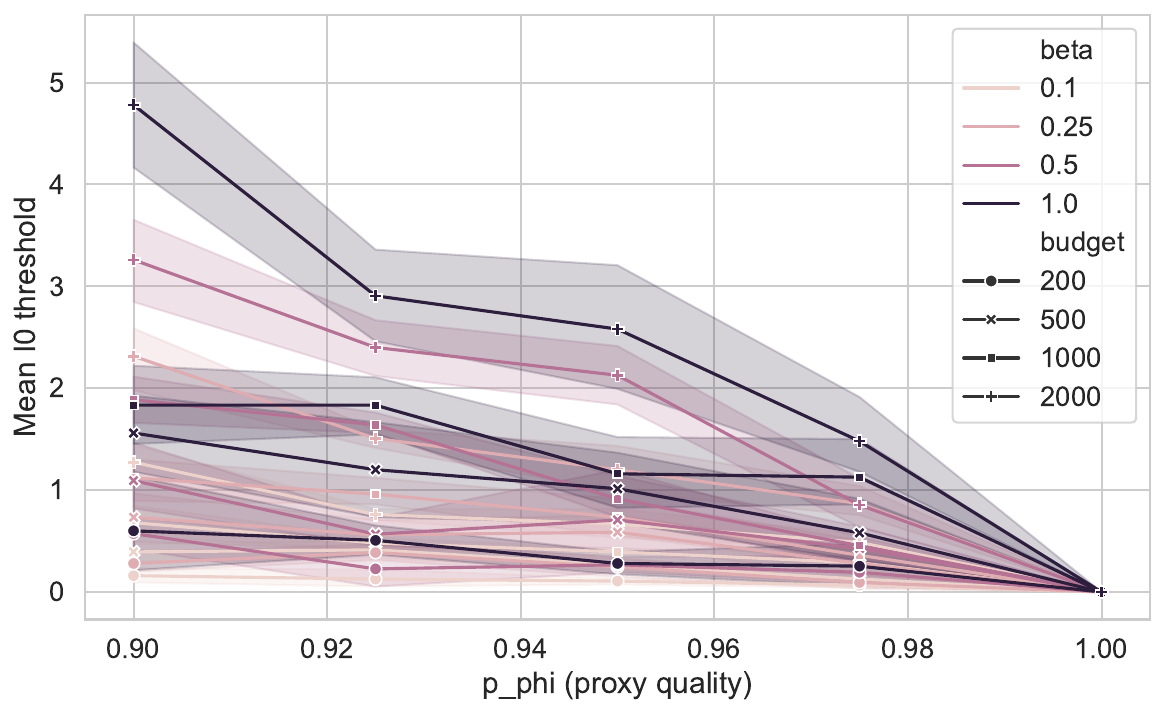}
    \caption{Threshold $I_0$ as a function of proxy quality $p_\phi$.}
    \label{fig:threshold-proxy}
\end{figure}

Figure~\ref{fig:threshold-proxy} reports the rejection threshold $I_0$ 
as a function of the proxy quality $p_\phi$. The curves correspond to several
values of multiple verification budgets.
The figure shows a clear monotonic structure: as $p_\phi$ increases, the threshold
drops sharply, reflecting the reduced noise floor of the proxy.

For low-quality proxies ($p_\phi \leq 92\%$), the threshold remains high, meaning
that only large manipulations can be detected. Once $p_\phi$ crosses a critical
region around $94\%$--$96\%$, the threshold collapses, enabling detection even
under modest budgets. This behaviour mirrors the phase transition predicted by
Theorem~2: the binomial noise floor becomes dominated by the manipulation
signal.

\subsection{Detection Probability as a Function of Proxy Quality}

\begin{figure}[h]
    \centering
    \includegraphics[width=.5\linewidth]{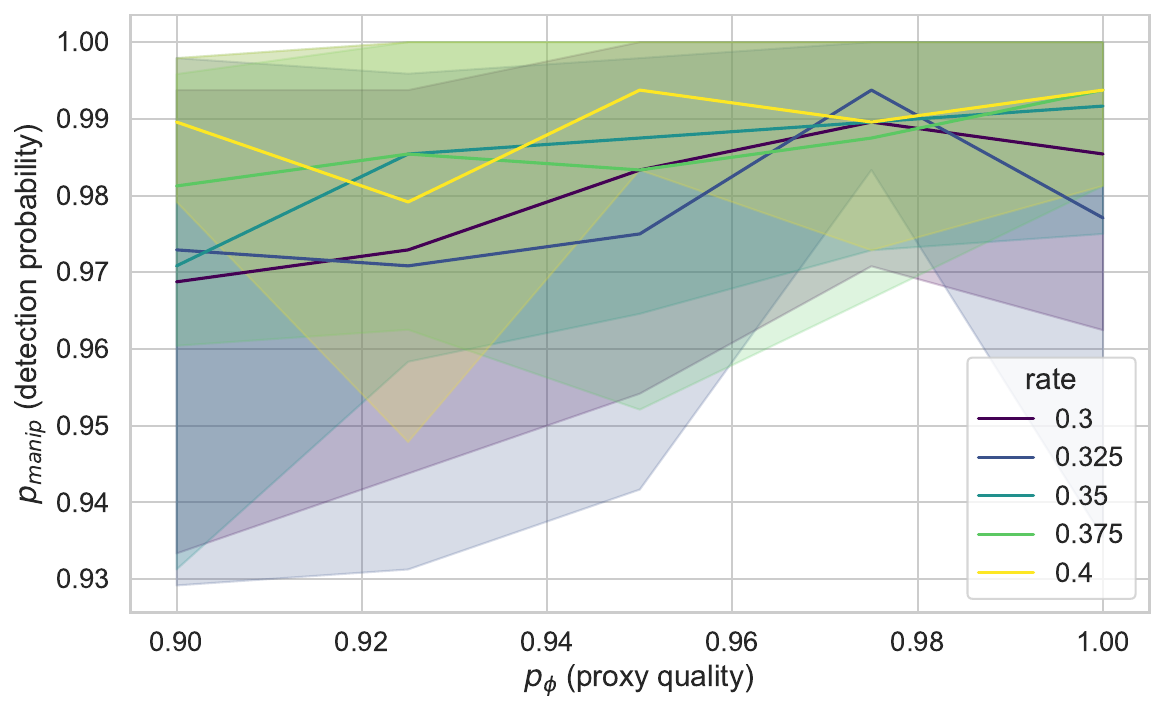}
    \caption{Detection probability $p_{manip}$ as a function of proxy quality $p_\phi$.}
    \label{fig:prob-proxy}
\end{figure}

Figure~\ref{fig:prob-proxy} displays the empirical detection probability as a
function of $p_\phi$ for several manipulation rates.

Detection probability remains far from the nominal Type~I error. Detection power increases rapidly, especially for manipulation rates
$\gamma \leq 35\%$. This confirms the theoretical prediction that detection power is
jointly monotone in $p_\phi$ and $\gamma$.

\subsection{Impact of Verification Budget}

\begin{figure}[h]
    \centering
    \includegraphics[width=.5\linewidth]{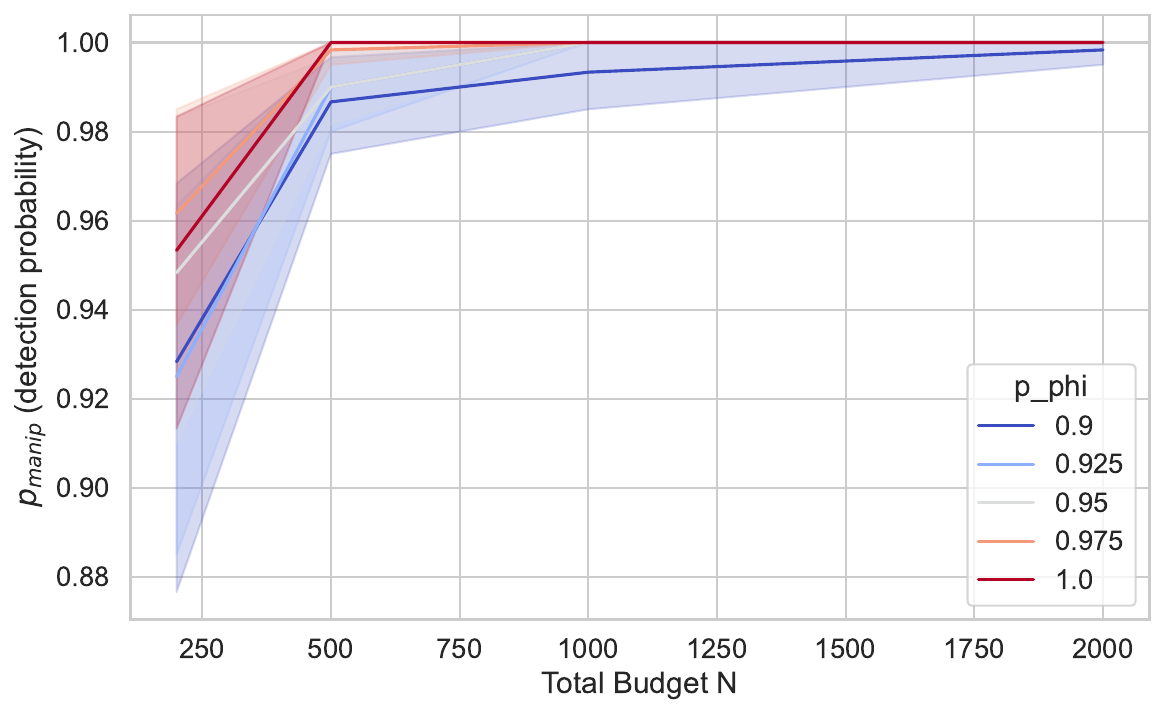}
    \caption{Impact of verification budget on detection probability.}
    \label{fig:budget-impact}
\end{figure}

Figure~\ref{fig:budget-impact} illustrates how the verification budget $N_{\rm verify}$
affects detection probability.

Increasing $N_{\rm verify}$ shifts the detection curves upward uniformly. The
effect is nonlinear: the first few hundred verification queries yield the largest
gains, while returns diminish beyond $N_{\rm verify} \approx 1000$. This aligns
with the binomial tail behaviour in Eq.~(6), where variance shrinks as
$1/\sqrt{N_{\rm verify}}$.

\subsection{Effect of the Audit/Verify Ratio $\beta$}

\begin{figure}[h]
    \centering
    \includegraphics[width=.5\linewidth]{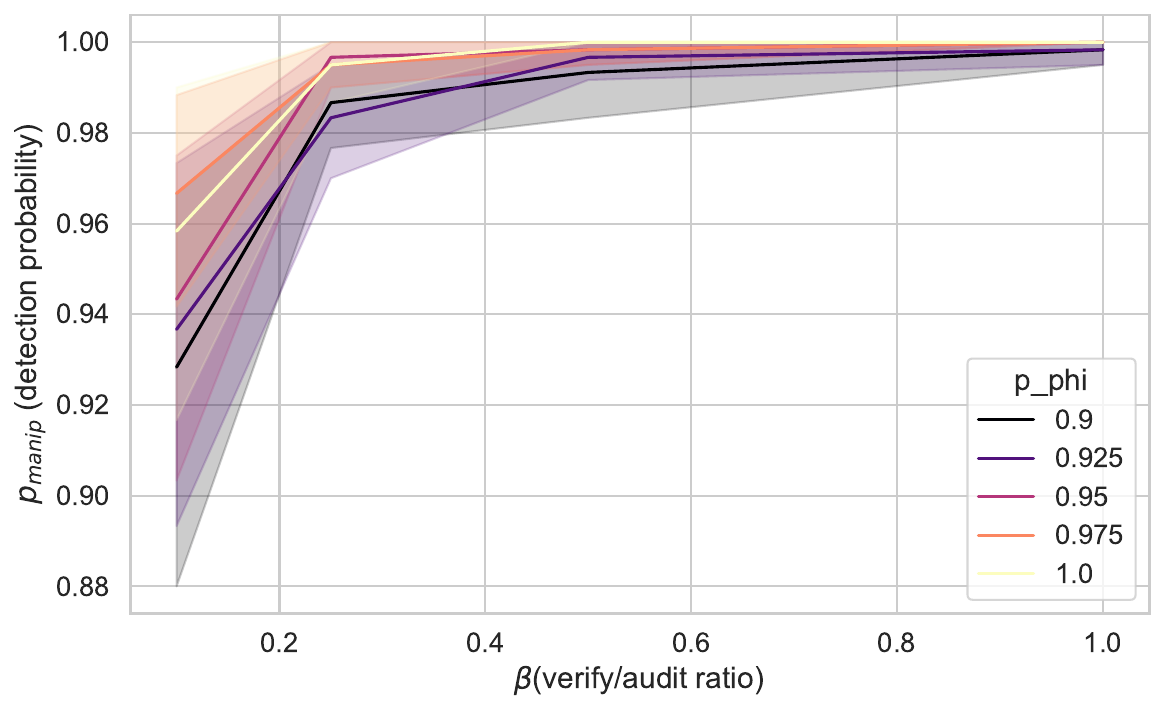}
    \caption{Impact of the audit/verify ratio $\beta$ on detection probability.}
    \label{fig:beta-impact}
\end{figure}

Figure~\ref{fig:beta-impact} shows how the ratio
$\beta = N_{\rm verify}/(N_{\rm audit} + N_{\rm verify})$
influences detection probability.

Higher values of $\beta$ systematically improve detection power. This reflects the
fact that the fairness test (based on $N_{\rm audit}$) becomes less informative
once the manipulation rate exceeds the single-source blind spot $\gamma_0$.
Allocating more budget to cross-verification directly strengthens the consistency
test, which is the only component capable of detecting fairwashing.

\subsection{Conclusion}

Across all figures, a coherent pattern emerges. Detection power displays a clear
\emph{phase transition} with respect to proxy quality: below a critical accuracy
threshold, the manipulation signal remains drowned in proxy noise, whereas above
it, detection becomes sharply more reliable. Increasing the verification budget
further amplifies this effect, though with diminishing returns as the binomial
variance stabilizes. The audit/verify ratio $\beta$ also plays a decisive role,
governing whether the auditor operates above or below the single-source blind
spot identified in Section~3.2. Taken together, these visualizations confirm the
theoretical predictions of Theorem~2 and Theorem~3 and reinforce the central
message of the paper: \emph{imperfect proxies are not merely usable—they are
sufficient to close the structural blind spot of single-source auditing, provided
their quality is characterized and the verification budget is allocated
appropriately.}

\section{Additional Experiments with Other Fairwashing Techniques}\label{app:fairwashing}

To assess the robustness of the 2SAM beyond the positive-discrimination strategy
considered in the main text, we evaluate three additional fairwashing techniques
on the UCI Adult dataset: Score Shifting, Targeted Flipping, and Randomized
Fairwashing. For each technique, we report in Table~\ref{tab:fairwashing} the
minimal manipulation rate $\gamma^\star$ required to reach a disparate impact
$\mathrm{DI} \geq \tau = 80\%$, along with the resulting disparate impact and
the empirical detection probability $p_{\mathrm{manip}}$ achieved by the 2SAM
using the linguistic proxy $\phi_{\mathrm{gen}}$ ($p_\phi = 94\%$).

\begin{table}[h]
\centering
\begin{tabular}{lccc}
\toprule
Technique & $\gamma^\star$ & $\mathrm{DI}(\gamma^\star)$ & $p_{\mathrm{manip}}$ \\
\midrule
Positive Discrimination & 32\% & 83\% & 70\% \\
Score Shifting          & 54\% & 82\% & 57\% \\
Targeted Flipping       & 30\% & 82\% & 37\% \\
Randomized              & 36\% & 81\% & 80\% \\
\bottomrule
\end{tabular}
\caption{Minimal manipulation level $\gamma^\star$ required to reach
$\mathrm{DI} \geq 80\%$ for each fairwashing technique, along with the
resulting disparate impact and empirical detection probability.}
\label{tab:fairwashing}
\end{table}

\paragraph{Positive Discrimination.}
This strategy consists in flipping a fraction $\gamma$ of negative predictions 
for individuals in the protected group. It is the fairwashing technique considered in the paper. Formally, for any individual $i$ with 
$s_i = 0$ and $\hat{y}_i = 0$, the platform sets $\hat{y}_i^{\mathrm{fw}} = 1$ 
with probability $\gamma$. This corresponds to increasing the acceptance rate of the protected group.

\paragraph{Score Shifting.}
In this strategy, the platform modifies the internal scores produced by the 
classifier. For each protected individual $i$, the score $s_i$ is replaced by 
$s_i^{\mathrm{fw}} = \min\{1,\, s_i + \gamma \Delta\}$, where $\Delta$ controls 
the maximum shift. Predictions are then obtained by thresholding 
$s_i^{\mathrm{fw}}$.

\paragraph{Targeted Flipping.}
This strategy selects the subset of protected individuals with negative 
predictions but highest classifier scores. Let $S = \{ i : s_i \text{ high}, 
s_i < \tau, s_i \in \text{protected group} \}$. The platform flips the top 
$\gamma |S|$ individuals in $S$. This maximizes the fairness improvement for a 
given manipulation budget, and corresponds to an adversarial optimization of 
the disparate impact.

\paragraph{Randomized Fairwashing.}
In this stochastic strategy, each protected individual with a negative 
prediction is flipped independently with probability $\gamma$. Formally, 
$\hat{y}_i^{\mathrm{fw}} = 1$ with probability $\gamma$ if 
$\hat{y}_i = 0$ and $s_i = 0$. This randomization reduces the correlation 
structure of inconsistencies and makes the manipulation more difficult to 
detect through binomial consistency tests.

The results confirm that the 2SAM generalises across fairwashing strategies,
though detection power varies with the structural properties of each technique.
Score Shifting requires a substantially larger manipulation rate ($\gamma^\star =
54\%$) to reach compliance, which produces a stronger inconsistency signal and
yields the second-highest detection power ($p_{\mathrm{manip}} = 57\%$).
Targeted Flipping, by contrast, concentrates manipulations on the
highest-scoring individuals in the protected group, minimising the number of
label flips needed while maximising the fairness gain; this efficiency comes at
a cost for the auditor, as the resulting inconsistency signal is weaker and
detection power drops to $37\%$. Randomized Fairwashing achieves the highest
detection probability ($80\%$) despite a moderate $\gamma^\star$: because flips
are applied independently at random, the inconsistencies are distributed across
the full protected group rather than concentrated on a specific subpopulation,
making the aggregate signal easier to detect through the binomial consistency
test. Taken together, these results show that the 2SAM is not specifically
tailored to positive discrimination: it exploits any label-level inconsistency
between the Audit API and the trusted stream, regardless of the mechanism
used to produce it.

\section{Additional experiments on COMPAS}\label{a:48611}

To extend the experimental evaluation of the 2SAM beyond the UCI Adult dataset, we replicate the auditing game on the COMPAS recidivism dataset~\cite{angwin2016}, a widely studied benchmark in algorithmic fairness. COMPAS is a risk assessment tool used in the US criminal justice system to predict recidivism, and has been the subject of significant scrutiny for racial bias. We use race (Caucasian vs.\ non-Caucasian) as the sensitive attribute, two-year recidivism as the binary outcome and the following features: 'age','priors count','juv fel count','juv misd count', 'juv other count','c charge degree','sex','age cat','is recid'.

\paragraph{Model.}
We replace the logistic regression of Section~4 with a \texttt{random forest}. This choice is deliberate: the high expressivity of random forests allows the model to overfit the structural correlations between race and recidivism present in the data, yielding a nearly null disparate impact ($\mu \approx 0$) on the production model. This constitutes a more extreme bias regime than the Adult dataset, and provides a complementary stress test for the 2SAM.

\begin{figure}[ht]
 \centering
 \includegraphics[width=0.6\linewidth]{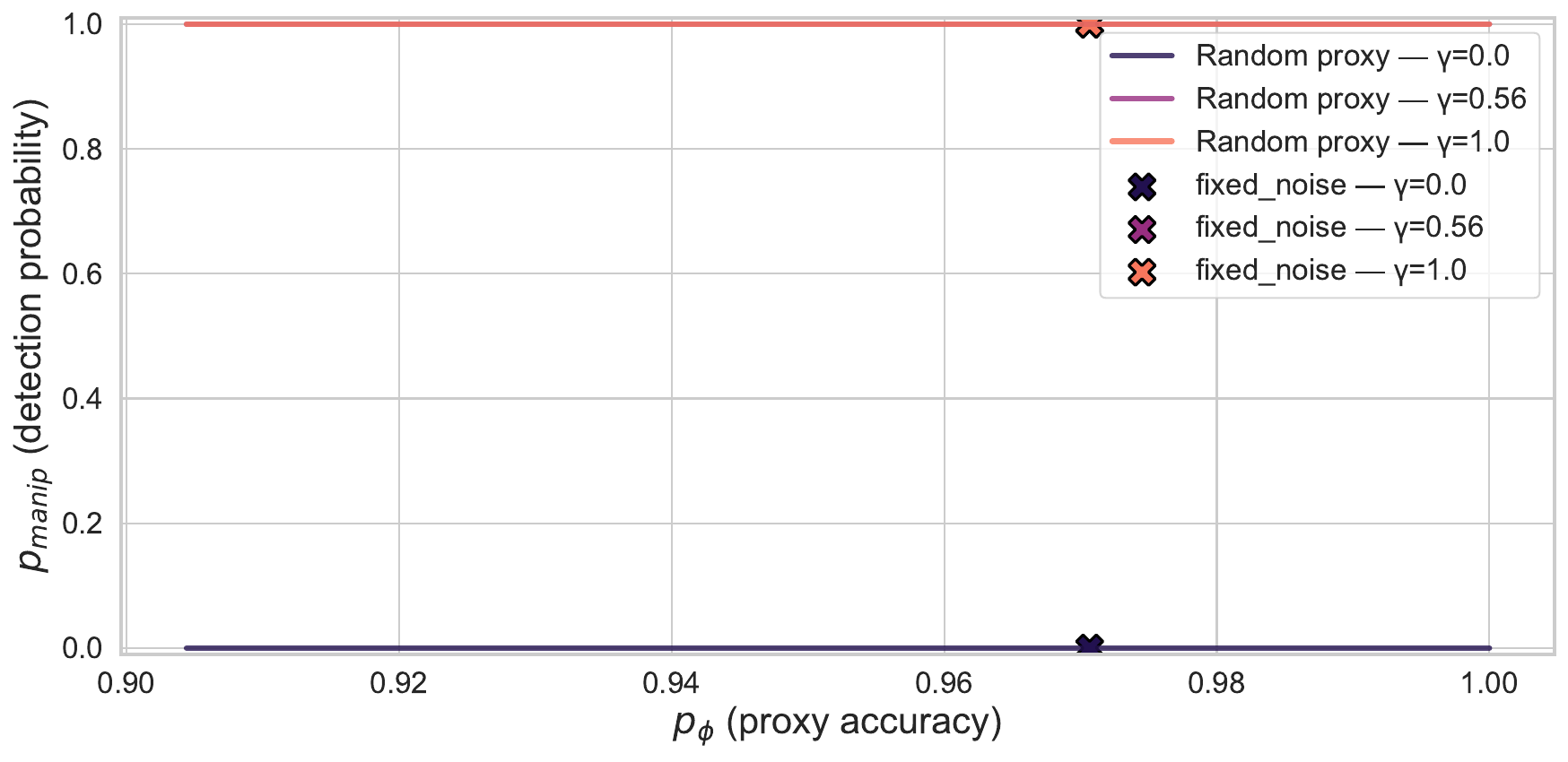}
 \caption{Detection power $p_\mathrm{manip}$ as a function of proxy quality $p_\phi$, for several manipulation rates $\gamma$.}
 \label{fig:proxy_quality:compas}
 \end{figure}

\paragraph{Effect of proxy quality on detection power (\Cref{fig:proxy_quality:compas}, analogue of Figure~1).}
The detection power curves are step linear in $p_\phi$, with negligible variance across runs. This regularity is a direct consequence of the near-zero disparate impact: since $\mu(h_{\mathrm{verify}}) \approx 0$, the platform must flip a large fraction of negative labels to pass the fairness threshold $\tau$, meaning that each manipulated query carries a strong inconsistency signal. This signal highlights the phase-transition structure observed on the Adult dataset between $p_\phi$ and $p_{\mathrm{manip}}$.

\begin{figure}[ht]
 \centering
 \begin{subfigure}[b]{0.48\linewidth}
 \centering
 \includegraphics[width=\linewidth]{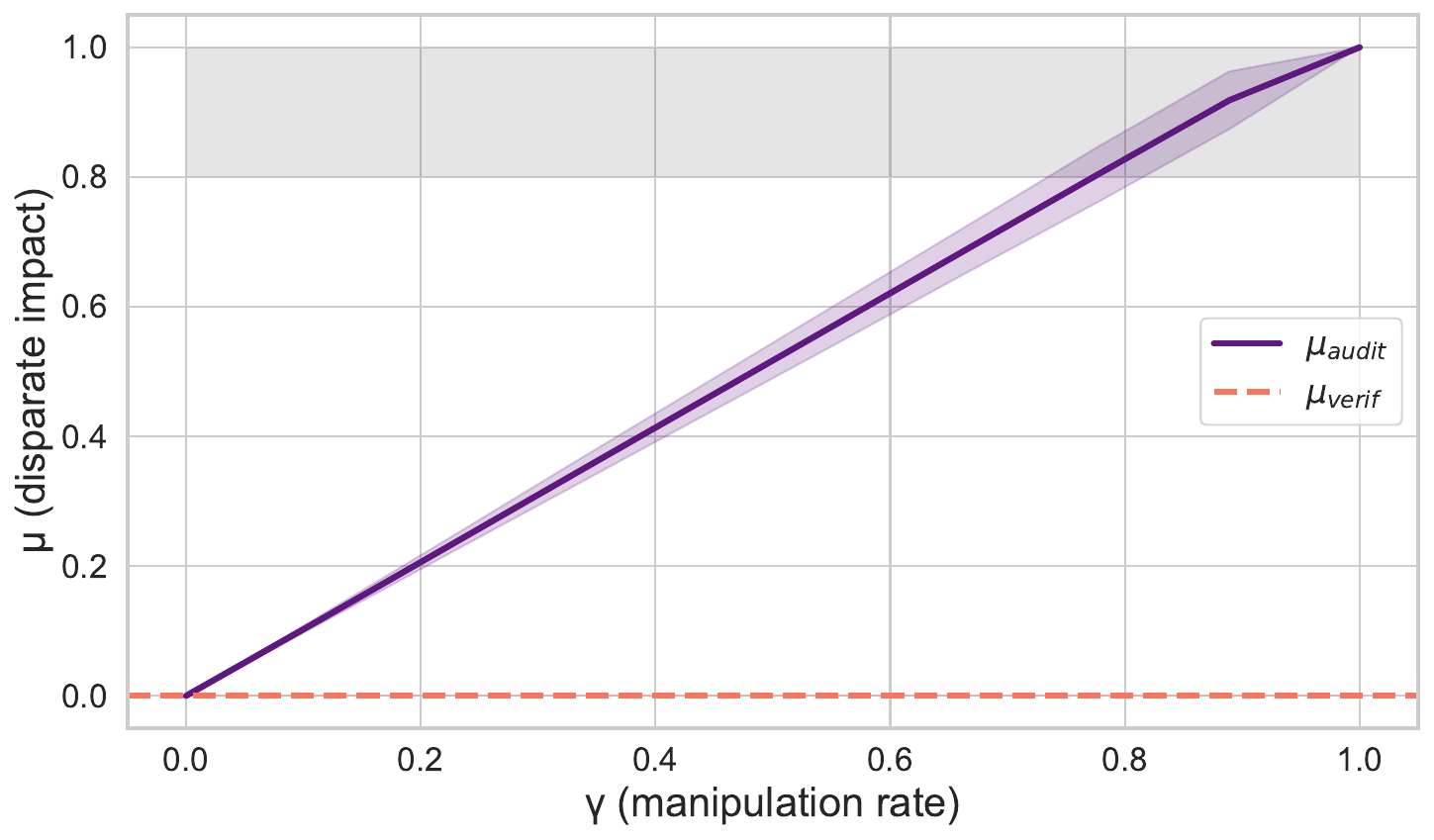}
 \caption{Estimated fairness as $\gamma$ increases. }
 \label{fig:lies_intermediate:compas}
 \end{subfigure}
 \hfill
 \begin{subfigure}[b]{0.48\linewidth}
 \centering
 \includegraphics[width=\linewidth]{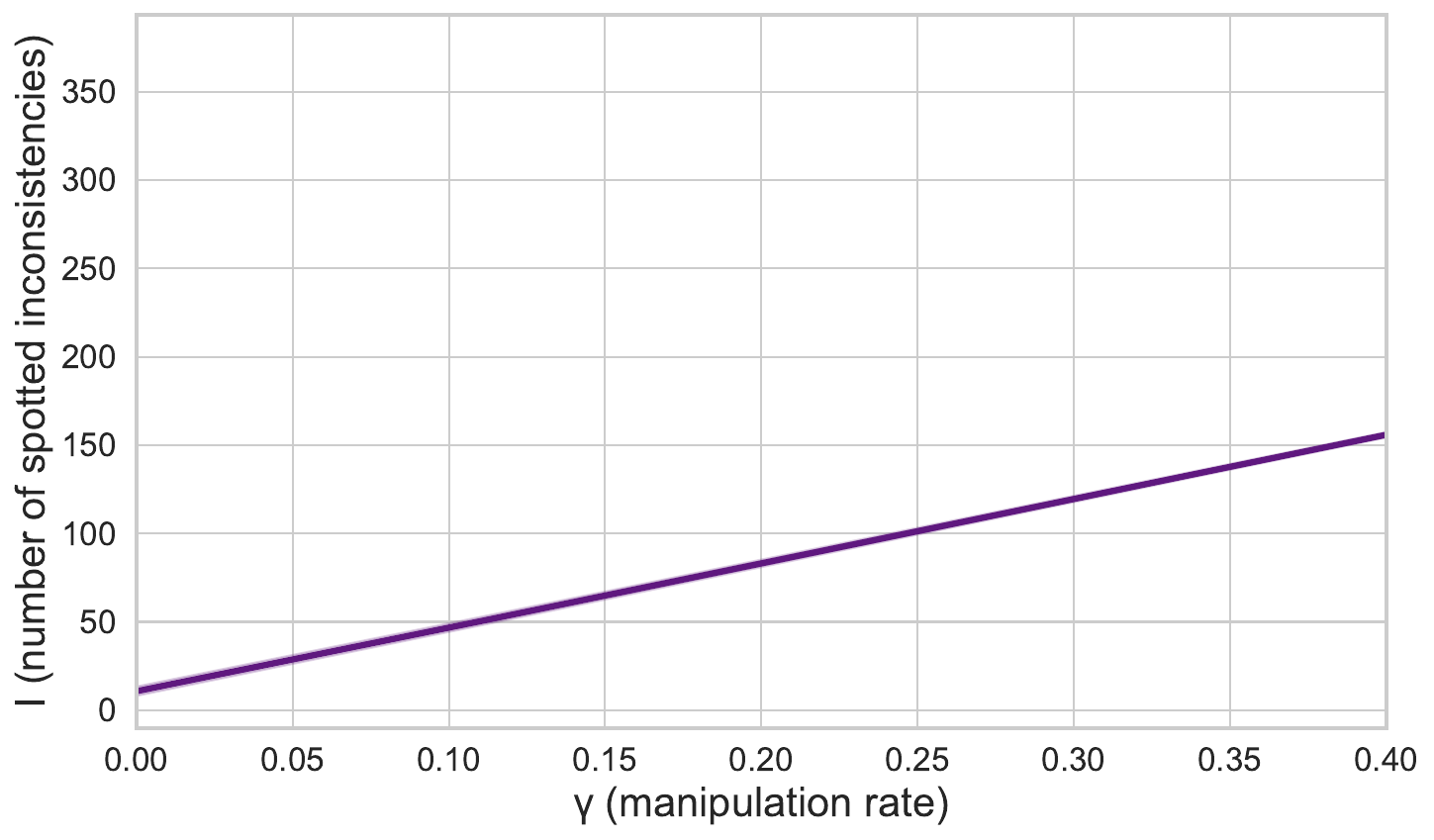}
 \caption{Inconsistency count $I$ as $\gamma$ increases.}
 \label{fig:inconsistency_count:compas}
 \end{subfigure}
 \caption{Single-source failure (left) and 2SAM detection (right) under increasing manipulation rate $\gamma$. The gray area marks the fair zone ($\hat{\mu} \geq \tau$). The single-source auditor incorrectly concludes compliance beyond $\gamma_0 \approx 32\%$.}
 \label{fig:detection:compas}
 \end{figure}

\paragraph{Single-source failure and 2SAM detection (\Cref{fig:detection:compas}, analogues of Figures~2a--2b with noisy proxy of fixed noise at $5\%$).}
The same structural pattern holds: as $\gamma$ increases, $\hat{\mu}(\manipulated)$ crosses $\tau$ and the single-source auditor incorrectly concludes compliance, while the inconsistency count $I$ rises monotonically. The absence of variance across runs reflects the same cause: the expressive Random Forest produces highly deterministic label-flipping patterns, reducing stochasticity in $I$ and confirming the robustness of Lemma~1 and Theorem~2 in this extreme bias regime.

\begin{figure}[h!]
 \centering
 \includegraphics[width=0.55\linewidth]{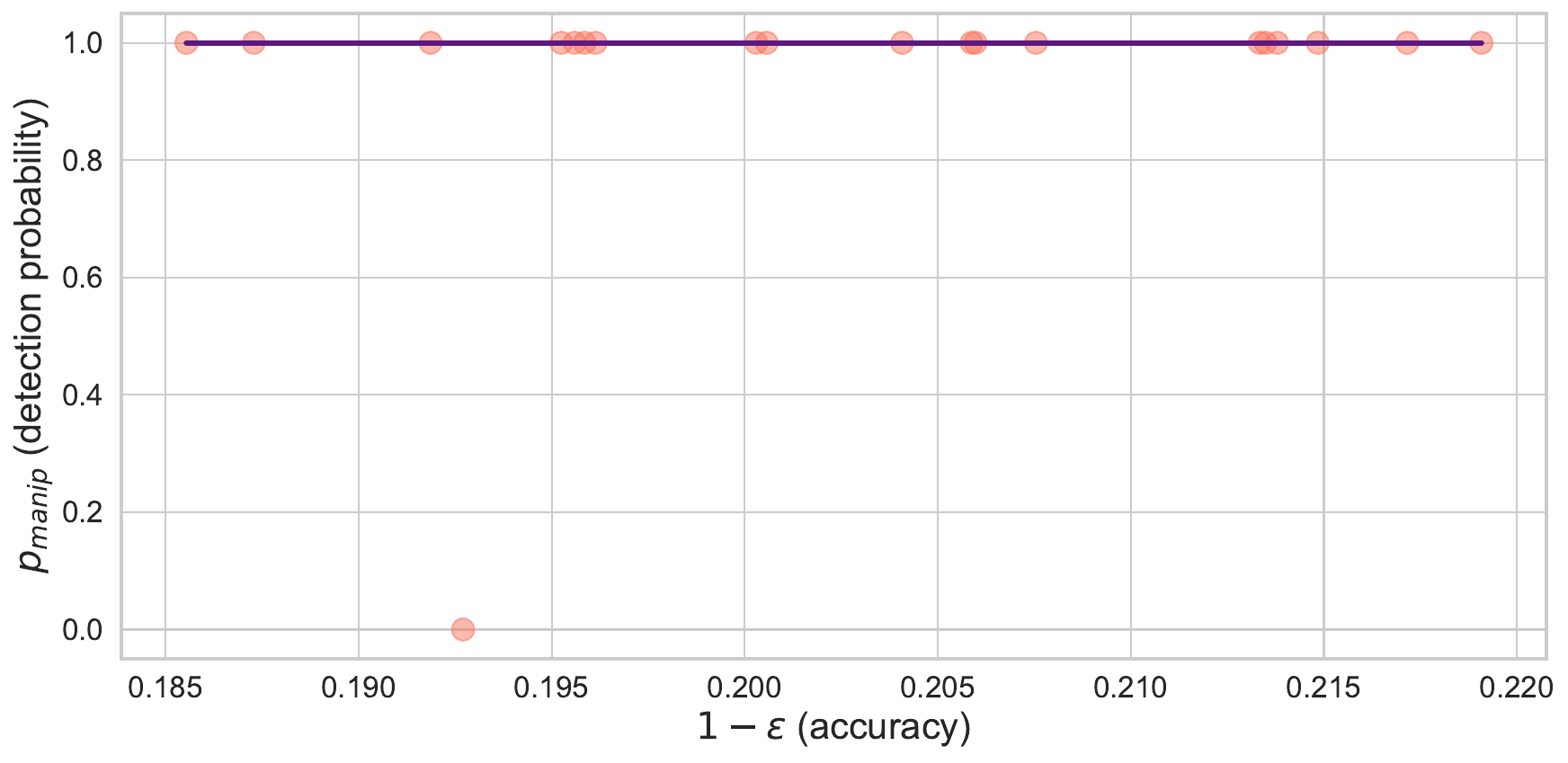}
 \caption{Pareto frontier for estimating Disparate Impact while checking for fairwashing under a fixed audit budget ($N = 750$). Each point corresponds to a budget allocation $\beta$.}
 \label{fig:pareto:compas}
 \end{figure}

\paragraph{Budget allocation (\Cref{fig:pareto:compas}, analogue of Figure~3).}
The Pareto frontier is qualitatively different from the Adult case. Because each individual manipulation is large---the platform must flip a substantial fraction of labels to move from $\mu \approx 0$ to $\hat{\mu}(\manipulated) \geq \tau$, the inconsistency signal is strong even at low verification budgets. As a result, the 2SAM achieves $p_{\mathrm{manip}} \approx 1$ across virtually all budget allocations $\beta$, while maintaining good estimation accuracy. This confirms Theorem~3: when $\gamma_0$ is large, the budget condition is easy to satisfy, and the trade-off between fairness estimation and manipulation detection nearly vanishes.

\paragraph{Takeaway.}
The COMPAS experiments confirm that the 2SAM is robust across bias regimes. When the production model is severely biased, the auditing task becomes easier rather than harder: larger required manipulations produce stronger inconsistency signals, allowing near-certain detection at negligible query cost. The Adult experiments remain the more demanding benchmark, as the logistic regression exhibits moderate bias that demands more careful budget allocation.

\section{Data for reproducibility}\label{app:repro}
The code will be open-sourced under the GPLv3 licence.

The experiments were performed on a machine equipped with an Apple M4 Pro processor (12 CPU cores) and 24 GB of memory.

Data is adapted from the Census Income dataset available in the UCI Machine Learning repository~\cite{Dua:2019}. This dataset is licensed under a Creative Commons Attribution 4.0 International (CC BY 4.0) license. This allows for the sharing and adaptation of the datasets for any purpose, provided that the appropriate credit is given.
The AdultIncome dataset contains $48,842$ instances ($32,072$ instances in the training set, $16,281$ in the test set, and $489$ in the audit set).

In addition to the characteristics of the AdultIncome dataset, we added random first names to all profiles following Top 10 names in France in 2024\cite{Insee}:
\par\textbf{Female names}: 'Louise', 'Jade', 'Ambre', 'Alba', 'Emma', 'Alma', 'Romy', 'Rose', 'Alice', 'Anna'.
\par \textbf{Male names}: 'Gabriel', 'Raphael', 'Louis', 'Leo', 'Noah', 'Arthur', 'Adam', 'Jules', 'Mael', 'Leon'

The logistic regression $f$ has an accuracy of $76 \%$.

\section{Disclosure of AI tools usage}
This manuscript benefited from the use of large language models during the writing process. In particular, Claude was used to assist with language editing of the text, and GitHub Copilot was used to revise the code. The authors reviewed and edited all generated suggestions and take full responsibility for the final content.

\end{document}